\documentclass[lettersize,journal]{IEEEtran}
\usepackage{amsmath,amsfonts}
\usepackage{algorithmic}
\usepackage{algorithm}
\usepackage{array}
\usepackage[caption=false,font=normalsize,labelfont=sf,textfont=sf]{subfig}
\usepackage{textcomp}
\usepackage{stfloats}
\usepackage{url}
\usepackage{verbatim}
\usepackage{graphicx}
\usepackage{cite}
\usepackage{fancyhdr}
\usepackage{multirow}
\usepackage{dsfont}

\usepackage{xspace}

\makeatletter
\DeclareRobustCommand\onedot{\futurelet\@let@token\@onedot}
\def\@onedot{\ifx\@let@token.\else.\null\fi\xspace}

\def\etal{\emph{et al}\onedot}
\makeatother

\usepackage{amsthm}
\newtheorem{myDef}{Definition}

\begin{document}

\title{Improving the Transferability of Adversarial Attacks on Face Recognition with Beneficial Perturbation Feature Augmentation}

\markboth{IEEE TRANSACTIONS ON COMPUTATIONAL SOCIAL SYSTEMS}{ZHOU \MakeLowercase{\textit{et al.}}: Improving the Transferability of Adversarial Attacks on Face Recognition with Beneficial Perturbation Feature Augmentation}

\author{Fengfan Zhou, Hefei Ling, Yuxuan Shi, Jiazhong Chen, Zongyi Li, Ping Li
}

\maketitle

\begin{abstract}
Face recognition (FR) models can be easily fooled by adversarial examples, which are crafted by adding imperceptible perturbations on benign face images. The existence of adversarial face examples poses a great threat to the security of society. In order to build a more sustainable digital nation, in this paper, we improve the transferability of adversarial face examples to expose more blind spots of existing FR models.
Though generating hard samples has shown its effectiveness in improving the generalization of models in training tasks, the effectiveness of utilizing this idea to improve the transferability of adversarial face examples remains unexplored.
To this end, based on the property of hard samples and the symmetry between training tasks and adversarial attack tasks, we propose the concept of hard models, which have similar effects as hard samples for adversarial attack tasks.
Utilizing the concept of hard models, we propose a novel attack method called \textbf{B}eneficial \textbf{P}erturbation \textbf{F}eature Augmentation \textbf{A}ttack (BPFA), which reduces the overfitting of adversarial examples to surrogate FR models by constantly generating new hard models to craft the adversarial examples.
Specifically, in the backpropagation, BPFA records the gradients on pre-selected feature maps and uses the gradient on the input image to craft the adversarial example. In the next forward propagation, BPFA leverages the recorded gradients to add beneficial perturbations on their corresponding feature maps to increase the loss. Extensive experiments demonstrate that BPFA can significantly boost the transferability of adversarial attacks on FR.
\end{abstract}


\section{Introduction}
As a disruptive technology of artificial intelligence, FR has been widely used in scenarios with high-security requirements in our society, such as airport security checks, financial payment, and mobile phone unlocking\cite{genap}. However, previous works have shown that FR is vulnerable to adversarial examples\cite{adv_makeup}\cite{adv_glass}. That the attacker can successfully fool an FR model by adding imperceptible adversarial perturbations on input images of the FR model\cite{dfanet} harms the collective wellbeing of the society.

In addition, the adversarial face examples crafted by the attacker are transferable across different FR models. The transferability of adversarial face examples enables the attacker to attack the victim FR model successfully without knowing any information about it\cite{genap}\cite{amt_gan}\cite{dfanet}.

The feasibility of black-box attacks on FR models poses a significant threat to the building process of the sustainable digital nation.
For example, a government agency of a digital nation is responsible for deploying FR models at the country's borders to identify and track potential security threats. An attacker generates adversarial face examples that can bypass these models, leading to a national security breach. The government agency is unable to detect the threat, leading to potential harm to citizens and safety issues in the digital nation.
There are three stakeholders: (1) Citizens: They are impacted by the potential threats posed by adversarial attacks, as their identities and personal information can be compromised. They are the end-users and beneficiaries of the FR models in the digital nation. (2) The government: They are responsible for deploying and maintaining the FR models in the digital nation. The government is accountable for ensuring the security and privacy of citizens' personal information and preventing unauthorized access. (3) Technology companies: They are instrumental in developing and implementing FR models in the digital nation. They are responsible for ensuring that these models are secure, reliable, and comply with legal standards under the governance within the digital nation.

All the stakeholders mentioned above can be significantly impacted by black-box adversarial attacks on FR models. Therefore, to better govern and ensure the sustainability of the digital nation, improving the transferability of the adversarial face examples to expose more blind spots of existing FR models is of extreme urgency.

In recent years, adversarial attacks on FR have made great progress\cite{adv_glass}\cite{adv_hat}\cite{effective_and_robust}. However, only a few works focus on the improvement of transferability of adversarial face examples\cite{dfanet}. We argue that the transferability of adversarial face examples is important because the transferability affects the black-box attack performance of adversarial face examples. There exists a large gap between the black-box and white-box performance of adversarial face examples crafted by existing adversarial attacks on FR. Therefore, the improvement of the transferability of adversarial face examples needs further research.

In most cases, the adversarial attack tasks are the optimization processes that the models are fixed while the inputs are optimized. In contrast, the training tasks are optimization processes that the model are optimized while the inputs are fixed. Therefore, there exists symmetry between the two kinds of tasks.
Lots of works have utilized the symmetry to improve the transferability of adversarial examples, particularly the attacks based on advanced optimization\cite{mim}\cite{ni_fgsm}.
In training tasks, generating hard samples for data augmentation brings great improvement to the generalization of the model\cite{teach_aug}. However, in adversarial attack tasks, there are few works that use this idea to augment the surrogate model to improve the transferability of adversarial face examples. Therefore, the effectiveness of generating hard samples for augmentation in improving the transferability of adversarial face examples needs further exploration. Based on the symmetry between the adversarial attack tasks and the training tasks, the hard samples of training tasks correspond to the hard models of the adversarial attack tasks.
Therefore, we attempt to generate hard models to augment the ``training set'' of the adversarial attack tasks. Hard samples are easy to understand. However, hard models are not. In general, the losses of hard samples are greater than the losses of normal samples for the same model.
Therefore, based on the property of hard samples, we define the hard models as the models whose losses are greater than the normal models for the same input. The definition of the hard models is in Definition \ref{Definition:hard_model}.

To increase the loss, we plan to add noises that can be pitted against the adversarial examples on the feature maps of the model to generate the hard models, and these noises are called beneficial perturbations\cite{Beneficial_Perturbation_Network}.

Adding the beneficial perturbations on the feature maps can increase the loss so as to generate harder models to augment the models of adversarial attacks. However, we need one more backpropagation in each iteration to calculate the beneficial perturbations on the feature maps, increasing the computational overhead. To address this issue, we use the gradients in the last backpropagation to simulate the gradients in this backpropagation to craft the beneficial perturbations. We name the proposed attack method as \textbf{B}eneficial \textbf{P}erturbation \textbf{F}eature Augmentation \textbf{A}ttack (BPFA).

Our main contributions are summarized as follows:
\begin{itemize}
	\item We propose a novel adversarial attack method on FR, named BPFA, to enhance the transferability of crafted adversarial face examples thereby facilitating the development of governance in nations moving to higher levels of digitalization. In the process of crafting adversarial examples using BPFA, the beneficial perturbations are added on the pre-selected feature maps of the FR model to constantly generate hard models to improve the transferability of the crafted adversarial examples.
	\item We explore the properties of beneficial perturbations added on the feature maps of the FR model. We find that the beneficial perturbations added on the feature maps may have some semantic information and analyze the causes for the semantic information.
	\item Extensive experiments show that our proposed BPFA can improve the black-box attack performance of existing adversarial attacks on FR without harming the white-box attack performance.
\end{itemize}

\section{Related Works}
\subsection{Beneficial Perturbation}
Beneficial perturbation was proposed by Wen and Itti\cite{Beneficial_Perturbation_Network}, and the original purpose of the beneficial perturbation is to improve the model's ability to resist adversarial examples (i.e., adversarial robustness\cite{DBLP:conf/cvpr/LiuCGLZS22}).
The optimization objective of beneficial perturbations is the same as the optimization objective of the task on which the perturbations are applied and is opposite to the optimization objective of adversarial perturbations. Therefore, the beneficial perturbations are not intended to fool the model and are \textit{beneficial} for the task on which the perturbations are applied.
The crating method of beneficial perturbation is similar to that of adversarial perturbation, which can be simply obtained by negating the optimization direction of the adversarial perturbations.
\subsection{Adversarial Attacks on Face Recognition}
As FR constitutes a crucial technique in the construction of a digital nation, adversarial attacks on FR represent a significant threat to its sustainability. Therefore, it is imperative that social innovation and governance are undertaken to further research and identify blind spots in the face of adversarial attacks on FR. Adversarial attacks on FR can be mainly classified into two categories: gradient-based attacks and generator-based attacks.
Most gradient-based adversarial attacks on FR are improved based on I-FGSM\cite{i_fgsm}. The FR belongs to the open-set task for which embedding-level loss is more suitable. Therefore, Zhong and Deng\cite{fim} proposed FIM. Compared with I-FGSM, the main improvement of FIM is to change the label-level loss to the embedding-level loss. To improve the transferability of adversarial face examples, Zhong and Deng\cite{dfanet} proposed DFANet. DFANet performs dropout on the feature maps of the convolutional layers during the forward propagation to obtain ensemble-like effects. For face encryption, Yang \etal\cite{tip_im} proposed TIP-IM. Different from FIM and DFANet, TIP-IM focuses on attacking face classification tasks. The main operation of TIP-IM is to use MMD\cite{mmd} loss to improve the visual quality of the crafted adversarial examples and use the greedy insertion to select the optimal victim image from a pre-defined gallery set.
The generator-based adversarial attacks on FR consist of two stages: training and inference. In the training stage, the training method of GAN\cite{gan} is used. In the inference stage, only the generator is used to generate adversarial examples.
Currently, the generator-based adversarial attacks on FR mainly focus on generating adversarial examples of a particular part of the face\cite{genap} and generating adversarial examples based on makeup transfer\cite{adv_makeup}\cite{amt_gan}.

In addition, adversarial attacks on FR can also be classified into impersonation (targeted) attacks and dodging (untargeted) attacks\cite{adv_makeup}. Impersonation attacks on FR aim to fool the FR model into recognizing adversarial examples as the face images of a pre-selected victim identity, while dodging attacks on FR aims to fool the FR model into recognizing adversarial examples as the face images of someone other than the attacker identity.
\subsection{Transferable Adversarial Attacks}
To build a more sustainable digital nation, it is crucial to explore the adversarial vulnerability of neural networks in the black-box setting. Therefore, various adversarial attacks with the aim of improving the transferability have been proposed\cite{ssa6}\cite{DBLP:conf/cvpr/ZhangWHHW0L22}.

As a direct method to improve the transferability, data augmentation has been well studied. Xie \etal \cite{dim} propose DI, DI applies random transformation on the input images in every iterations. Lin \etal \cite{ni_fgsm} propose SI, SI further adapts scale transformation operation to the input and uses the gradients of the loss with respect to the scaled copies to update the adversarial examples. Wang \etal \cite{admix} proposed Admix, Admix utilizes the gradients of input image admixed with a small portion of other images to craft adversarial examples to better improve the transferability.
Long \etal \cite{ssa6} proposed SSA, SSA applies a spectrum transformation to the input and augments the input in the frequency domain. 

Advanced optimization is also an effective strategy. Dong \etal \cite{mim} propose MI, MI adapts the momentum term to facilitate the adversarial examples to escape from the poor local optimum of the surrogate model. Lin \etal \cite{ni_fgsm} propose NI, NI utilizes the Nesterov accelerated gradient to look ahead to further improve the transferability.
Wang \etal \cite{vt} propose VT, VT adapts the gradient variance to stabilize the optimization direction. 
\section{Methodology}
This section presents our proposed BPFA method. Section \ref{problem_formulation} introduces the problem formulation. Section \ref{sec:bpfa_detailed_construction} focuses on the detailed construction of BPFA. To provide a clearer understanding of BPFA, Section \ref{sec:crafting_method_of_bp} elaborates the crafting method of beneficial perturbations added on feature maps, and Section \ref{sec:record_grad_w_mb} elaborates the computational saving method of BPFA.
\subsection{Problem Formulation}\label{problem_formulation}
Let $\mathcal{F}^{vct}(x)$ be an FR model used by government (a.k.a, victim FR model in our setting) to extract the embedding vector from a citizen’s face image $x$. Let $x^{s}$ be the image of the attacker that aims to attack the FR system in the digital nation and $x^t$ be the victim image of the victim citizen that the attacker aims to attack.
The objective of the impersonation attacks of FR model in the digital nation is to fool $\mathcal{F}^{vct}$ to recognize $x^{adv}$ as $x^t$ in the constraint that $x^{adv}$ is similar to $x^{s}$ visually. Specifically, the objective can be expressed as:
\begin{equation}
	\begin{gathered}
		x^{adv}=\mathop{\arg\min}\limits_{x^{adv}}\left(\mathcal{D}\left(\mathcal{F}^{vct}\left(x^{adv}\right), \mathcal{F}^{vct}\left(x^{t}\right)\right)\right) \\
		\text{s.t.} \Vert x^{adv} - x^{s}\Vert_p \leq \epsilon
		\label{eq:opt_obj_of_impersonation}
	\end{gathered}
\end{equation}
where $\mathcal{D}$ is a pre-selected distance metric, and $\epsilon$ is the maximum allowable perturbation magnitude.
In contrast, the objective of the dodging attacks of the FR model in the digital nation is to facilitate $\mathcal{F}^{vct}(x)$ cannot recognize $x^{adv}$ as $x^s$ in the constraint that $x^{adv}$ is similar to $x^{s}$ visually.
Similar to the impersonation attacks, the objective of the dodging attacks can be expressed as:
\begin{equation}
	\begin{gathered}
		x^{adv}=\mathop{\arg\max}\limits_{x^{adv}}\left(\mathcal{D}\left(\mathcal{F}^{vct}\left(x^{adv}\right), \mathcal{F}^{vct}\left(x^{s}\right)\right)\right) \\
		\text{s.t.} \Vert x^{adv} - x^{s}\Vert_p \leq \epsilon
		\label{eq:opt_obj_of_dodging}
	\end{gathered}
\end{equation}

\subsection{Beneficial Perturbation Feature Augmentation Attack}\label{sec:bpfa_detailed_construction}
In the digital nation, the deployed FR models is typically kept under tight wraps by governments or other stakeholders. The attacker cannot obtain the victim model $\mathcal{F}^{vct}$.
Therefore the optimization objective of Eq. \ref{eq:opt_obj_of_impersonation} and Eq. \ref{eq:opt_obj_of_dodging} cannot be directly realized. A common method to realize the optimization objective of Eq. \ref{eq:opt_obj_of_impersonation} and Eq. \ref{eq:opt_obj_of_dodging} is to use a surrogate model $\mathcal{F}$ that the attacker can obtain to craft adversarial examples and transfer the crafted adversarial examples to the victim model to carry out the attack\cite{naseer2023boosting}\cite{DBLP:conf/eccv/YuanZS22}. This requires that the adversarial examples crafted using the surrogate model have strong transferability. Therefore, transferability is important for adversarial attacks.

In the training tasks other than crafting the adversarial examples, generating hard samples for augmentation has demonstrated its effectiveness in improving the generalization of models\cite{adv_aug}\cite{teach_aug}. Therefore, based on the symmetry and the effectiveness of hard samples, we plan to generate hard models for augmentation to improve the transferability of adversarial face examples. To generate hard models, we should research the property of hard samples and clarify the definition of hard models by the property of hard samples first.
In FR, the hard samples are the samples that are prone to mislead the FR model (e.g., a negative face pair from identical twins and a positive face pair from a single identity with a large age gap). These samples are often closer to the decision surface than normal samples. Generating hard samples to train the FR model can smooth the decision surface of the model, thereby improving the generalization of the model. Notwithstanding the various types of hard samples, the hard samples have a common property that the losses of them are greater than the losses of normal samples.
Therefore, we assume that the losses of the hard models are also greater than the normal models for the same input.
To make the explanation of hard models more clear, here we give the definition of hard model:
\begin{myDef}\label{Definition:hard_model}
	\textbf{Hard Model}\\Let $\mathcal{F}$ be a normal pre-trained FR model, $x$ and $x'$ be two face images. An FR model $\mathcal{H}$ that satisfies the following conditions is called a hard model that is relative to $\mathcal{F}$ and is called a hard model for short:

$\mathcal{L}\left(\mathcal{F}\left(x\right),\mathcal{F}\left(x'\right)\right)<\mathcal{L}\left(\mathcal{F}\left(x\right),\mathcal{H}\left(x'\right)\right)$

where $\mathcal{L}\left(e,e'\right)$ is the loss function to measure the distance of two extracted face embeddings $e$ and $e'$.
\end{myDef}

\begin{figure}[htbp]
	\centering
	\includegraphics[width=80mm]{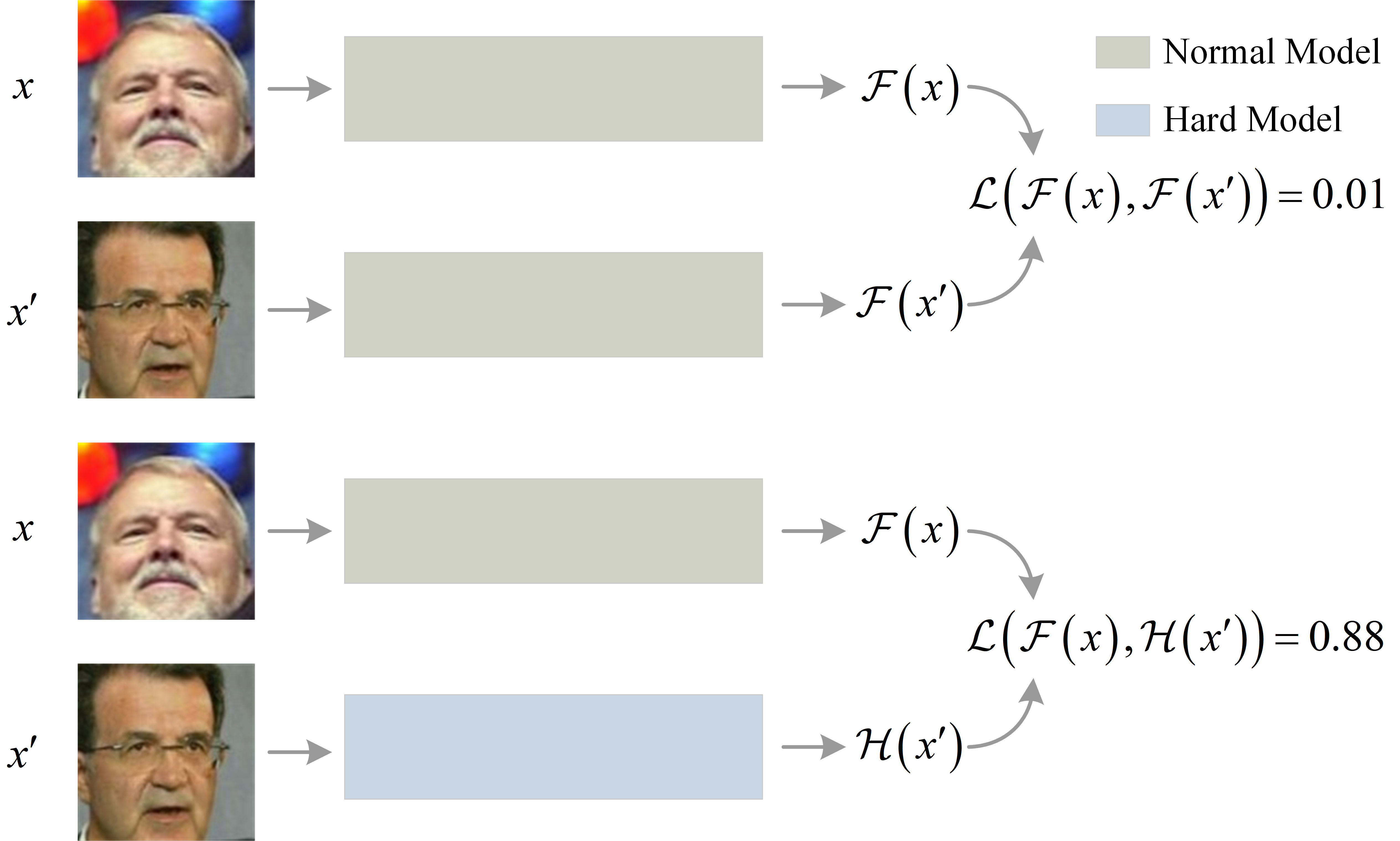}
	\caption{Illustration of the comparison between normal models and hard models. $x$ and $x^{\prime}$ are two face images. $\mathcal{F}$ and $\mathcal{H}$ are the normal model and hard model, respectively.}
	\label{fig:hard_model_compare}
\end{figure}

The examples of the normal model and hard model are shown in Fig. \ref{fig:hard_model_compare}.

We plan to generate hard models to augment the surrogate models, thereby improving the transferability of adversarial face examples.
To generate hard models, it is crucial to increase the loss of the model with respect to the same input. Therefore, we plan to add the beneficial perturbations which have an opposite optimization direction with the adversarial perturbations on the feature maps of the model to increase the loss.
In each iteration, the beneficial perturbations added on the feature maps are generally different, similar to using diverse FR models to craft the adversarial face examples. Our process of crafting adversarial examples by constantly generating new hard models can be seen as simulating the process of generating hard models for ensemble. We name the proposed attack method as \textbf{B}eneficial \textbf{P}erturbation \textbf{F}eature Augmentation \textbf{A}ttack (BPFA). 

The optimization objective of BPFA can be express as:
\begin{equation}
	\mathop{\min}\limits_{\Delta x^{adv}}\mathop{\max}\limits_{\Delta \Omega^{ben}}\mathcal{L}
\end{equation}
where $\Delta x^{adv}$ is the crafted adversarial perturbation that needs to be added on the attacker image $x^{s}$, and $\Delta \Omega^{ben}$ is the set of the crafted beneficial perturbations that need to be added on a pre-selected feature map set $\Omega$ of the surrogate model $\mathcal{F}$ to generate the hard model $\mathcal{H}$. $\mathcal{L}$ is the loss function that is different between the impersonation attacks and dodging attacks.

For impersonation attacks, the loss function can be expressed as:
\begin{equation}
	\mathcal{L}^{i}=\Vert\phi\left(\mathcal{H}\left(x^{adv}\right)\right)-\phi\left(\mathcal{F}\left(x^{t}\right)\right)\Vert^2_2\label{eq:bpfa_loss}
\end{equation}
where $\phi(x)$ denotes the operation that normalizes $x$. $x^{adv}$ is the adversarial example that initiates with the same value of $x^{s}$.

For dodging attacks, the loss function can be expressed as:
\begin{equation}
	\mathcal{L}^{d}=-\Vert\phi\left(\mathcal{H}\left(x^{adv}\right)\right)-\phi\left(\mathcal{F}\left(x^{s}\right)\right)\Vert^2_2\label{eq:bpfa_dodging_loss}
\end{equation}

Since the loss functions are the main difference between the optimization processes of impersonation attacks and dodging attacks, we will use $\mathcal{L}$ to denote both $\mathcal{L}^{i}$ and $\mathcal{L}^{d}$ to introduce our proposed method without ambiguity in the following.

The optimization process of BPFA corresponds to a minimax two-player game. Beneficial perturbations added on the feature maps of the model tend to increase the loss, while adversarial perturbations added on the input images tend to decrease the loss.
In particular, we use the following formula that minimizes the loss to craft adversarial examples\cite{i_fgsm}\cite{fim}:
\begin{equation}
	x^{adv}_t = \prod \limits_{x^s, \epsilon}\left( x^{adv}_{t-1}-\beta {\rm sign}\left(\nabla_{x^{adv}_{t-1}}\mathcal{L}\right)\right)\label{eq:craft_ax}
\end{equation}
where $x^{adv}_t$ is the adversarial example at the $t$-th iteration, $\beta$ is the step size to craft adversarial perturbations, and the $\prod$ is the clip operation which limits the pixel value of the crafted adversarial examples in $\left[x^s-\epsilon, x^s+\epsilon\right]$.

\subsection{Crafting Method of Beneficial Perturbation Added on Feature Maps}\label{sec:crafting_method_of_bp}

As a typical application of minimax, GANs\cite{gan} are not easy to optimize. We argue that an important reason why GANs are not easy to optimize is that the generators and discriminators do not converge at the same time easily. To prevent BPFA from having the same problem as GANs, we only craft one-step, one-time beneficial perturbations (i.e., all beneficial perturbations are crafted by one-step attacks and are deleted after they are used). Therefore, we need not guarantee the convergence of beneficial perturbations and only need to guarantee the convergence of the adversarial examples, reducing the optimization difficulty of BPFA.

In the following, we will introduce the crafting method of beneficial perturbations added on feature maps in detail.
Let $f_i$ be the $i$-th layer in $\mathcal{F}$. Let $\mathcal{F}_{i,j}$ be the segment of $\mathcal{F}$ from $f_i$ to $f_j$ which can be denoted as:
\begin{equation}
	\mathcal{F}_{i,j}=f_i \circ f_{i+1} \circ f_{i+2} \circ \cdots f_j\label{eq:f_i_j}
\end{equation}
where $\circ$ denote the operation of composition of function:
\begin{equation}
	\left(f_i \circ f_{i+1}\right)\left(x\right)=f_{i+1}\left(f_i\left(x\right)\right)
\end{equation}

Therefore, $\mathcal{F}$ can be denoted as:
\begin{equation}
	\mathcal{F}=\mathcal{F}_{1,n}\label{eq:mathcal_f}
\end{equation}
where $n$ is the number of layers of $\mathcal{F}$.

If we use $\mathcal{L}$ to craft adversarial examples, we can use the following formula\cite{fgsm}:
\begin{equation}
	x^{adv}=x^{s}-\eta {\rm sign}\left(\nabla_{x^{s}}\mathcal{L}\right)\label{eq:craft_ap_on_input}
\end{equation}
where $\eta$ is a nonnegative step size. A larger $\eta$ indicates a larger magnitude of added perturbations. Let the feature map in the $i$-th layer be $\omega$, which can be denoted as:
\begin{equation}
	\omega=\mathcal{F}_{1,i}\left(x^{adv}\right)\label{eq:omega_def}
\end{equation}

Let the index set of pre-selected layers to be added beneficial perturbations be $\Psi$. Therefore, the pre-selected feature map set $\Omega$ can be expressed as:
\begin{equation}
	\Omega=\{\mathcal{F}_{1,i}\left(x^{adv}\right): i \in \Psi\}\label{eq:omega_psi_relation}
\end{equation}

The formula for adding an adversarial perturbation $\Delta\omega^{adv}=\omega^{adv}-\omega$ to $\omega$ can be denoted as:
\begin{equation}
	\omega^{adv}=\omega-\eta {\rm sign}\left(\nabla_{\omega}\mathcal{L}\right)\label{eq:craft_ap_on_feature}
\end{equation}

Eq. \ref{eq:craft_ap_on_feature} is improved based on FGSM\cite{fgsm}, and the linearity assumption of FGSM is still applicable to Eq. \ref{eq:craft_ap_on_feature}. Based on the linearity assumption of the model, the objective of the adversarial perturbation crafted by Eq. \ref{eq:craft_ap_on_feature} is to minimize the loss $\mathcal{L}$. 

Therefore, if we want to craft a beneficial perturbation $\Delta\omega^{ben}=\omega^{ben}-\omega$ that maximizes the loss, we can change the loss function simply by negating $\mathcal{L}$. Combined with the fact that the ${\rm sign}$ function is an odd function, we can conclude the formula for calculating $\omega^{ben}$:

\begin{equation}
	\begin{aligned}
		\omega^{ben}&=\omega-\eta {\rm sign}\left(\nabla_{\omega}\left(-\mathcal{L}\right)\right)\\
		&=\omega-\eta {\rm sign}\left(-\nabla_{\omega}\mathcal{L}\right)\\
		&=\omega+\eta {\rm sign}\left(\nabla_{\omega}\mathcal{L}\right)
	\end{aligned}\label{eq:craft_bp_on_feature}
\end{equation}

The main difference between Eq. \ref{eq:craft_bp_on_feature} and Eq. \ref{eq:craft_ap_on_feature} is that Eq. \ref{eq:craft_bp_on_feature} has a positive sign before $\eta$, which is the opposite to the sign before $\eta$ in Eq. \ref{eq:craft_ap_on_feature}.

What we have described above is the method of adding a beneficial perturbation on a single feature map. In real cases, we can add beneficial perturbations on multiple feature maps. The process of adding beneficial perturbations on multiple feature maps can be viewed as the combination of repeatedly executing the process of adding a beneficial perturbation on a single feature map. In practice, we can use backpropagation of computational graphs to calculate the gradients on all the feature maps in the network at once and pick the gradients we need to craft beneficial perturbations.
The process can be easily implemented with existing deep learning frameworks by adding hooks in the layers that need to be added beneficial perturbations on to record the gradients of loss w.r.t the layers and use the gradients to calculate the beneficial perturbations using Eq. \ref{eq:craft_bp_on_feature}.

\subsection{Save Computational Overhead with Simulated Gradients}\label{sec:record_grad_w_mb}
If we want to add beneficial perturbations on feature maps to augment the model, we need to add the pre-computed beneficial perturbation $\Delta \omega^{ben}$ on the feature map $\omega$ of the model during the forward propagation. However, we cannot compute the gradients on feature maps before the backpropagation. Therefore, we cannot use Eq. \ref{eq:craft_bp_on_feature} to compute the beneficial perturbation that should be added on the feature map $\omega$ directly. If we use one backpropagation to compute the beneficial perturbations and another backpropagation to compute the adversarial perturbation, the computational overhead to craft adversarial examples will greatly increase. Inspired by the memory bank\cite{memory_bank} in the unsupervised feature learning\cite{moco}\cite{sim_clr} and free adversarial training\cite{free_at}, we record the gradients computed during the last backpropagation and use them to calculate the beneficial perturbations we need to add on feature maps to reduce the computational overhead.
The experimental results show that the beneficial perturbations that we calculated using the simulated gradients can still increase the loss well.

The overall algorithm of BPFA for impersonation attacks is summarized in Algorithm \ref{alg:bpfa}. We can change Algorithm \ref{alg:bpfa} to the algorithm for dodging attacks by replacing the 12th line with the following formula:
\begin{equation}
	\mathcal{L}_{t}=-\Vert\phi\left(\mathcal{F}_{\Psi_\theta, n}\left(\omega_{t,\theta}\right)\right)-\phi\left(\mathcal{F}\left(x^{s}\right)\right)\Vert^2_2\label{eq:dodging_loss_algorithm}
\end{equation}
and taking off the input restriction that the input face images of the algorithm should be negative face pairs.
\begin{algorithm}
	\renewcommand{\algorithmicrequire}{\textbf{Input:}}
	\renewcommand{\algorithmicensure}{\textbf{Output:}}
	\caption{Beneficial Perturbation Feature Augmentation Attack for Impersonation Attacks} 
	\label{alg:bpfa} 
	\begin{algorithmic}[1]
		\REQUIRE Negative face image pair $\{x^s, x^t\}$, the index set of pre-selected layers to be added beneficial perturbations $\Psi$, the step size of the beneficial perturbations $\eta$, the step size of the adversarial perturbations $\beta$, the maximum number of iterations $N_{max}$, maximum allowable perturbation magnitude $\epsilon$, the surrogate FR model $\mathcal{F}$. 
		\ENSURE An adversarial face example $x^{adv}_{N_{max}}$
		\STATE $x^{adv}_0=x^{s}$, $\theta=\vert\Psi\vert$, $s_1=1$
		\FOR{$t=1 ,..., N_{max}$}
		\STATE $\omega_{t,0}=x^{adv}$
		\FOR{$i=1 ,..., \theta$}
		\STATE $s_2=\Psi_i$
		\STATE $\omega_{t,i}=\mathcal{F}_{s_1,s_2}\left(\omega_{t,i-1}\right)$
		\STATE $s_1=s_2$
		\IF{$t \neq 1$}
		\STATE $\omega_{t,i}=\omega_{t,i}+\eta {\rm sign}\left(\nabla_{\omega_{t-1,i}}\mathcal{L}_{t-1}\right)$
		\ENDIF
		\ENDFOR
		\STATE $\mathcal{L}_{t}=\Vert\phi\left(\mathcal{F}_{\Psi_\theta, n}\left(\omega_{t,\theta}\right)\right)-\phi\left(\mathcal{F}\left(x^{t}\right)\right)\Vert^2_2$
		\STATE $x^{adv}_{t} = \prod \limits_{x^s, \epsilon}\left(x^{adv}_{t-1}-\beta {\rm sign}\left(\nabla_{x^{adv}_{t-1}}\mathcal{L}_{t}\right)\right)$ 
		\ENDFOR
	\end{algorithmic} 
\end{algorithm}

\begin{table*}[ht]
	\centering
	\small
	\caption{
		The attack success rates of impersonation attacks on LFW with the state-of-the-art attacks as the baseline. The first column represents the attacker models. The third to sixth columns in the first row represent the victim models. $*$ indicates white-box attacks.}
	\label{tab:asr_single_attack_lfw}
	\begin{tabular}{c|c|c|c|c|c}
		\hline
		& Attack         & \multicolumn{1}{c|}{FaceNet}      & \multicolumn{1}{c|}{MF}           & \multicolumn{1}{c|}{IRSE50}       & \multicolumn{1}{c}{IR152}         \\ \hline
		\multirow{5}{*}{FaceNet} & FIM / +BPFA    & \textbf{100.0*} / \textbf{100.0*} & 4.9 / \textbf{13.1}               & 11.9 / \textbf{26.2}              & 7.2 / \textbf{13.5}               \\
		& MI / +BPFA     & \textbf{100.0*} / \textbf{100.0*} & 7.7 / \textbf{19.0}               & 17.5 / \textbf{39.6}              & 13.1 / \textbf{25.1}              \\
		& DI / +BPFA     & \textbf{100.0*} / \textbf{100.0*} & 18.2 / \textbf{23.6}              & 30.7 / \textbf{40.4}              & 17.2 / \textbf{28.0}              \\
		& DFANet / +BPFA & \textbf{100.0*} / \textbf{100.0*} & 9.8 / \textbf{14.5}               & 19.5 / \textbf{28.9}              & 9.0 / \textbf{16.4}               \\
		& SSA / +BPFA    & \textbf{100.0*} / \textbf{100.0*} & 30.3 / \textbf{35.2}              & 43.3 / \textbf{49.2}              & 19.2 / \textbf{23.8}              \\ \hline
		\multirow{5}{*}{MF}      & FIM / +BPFA    & 6.7 / \textbf{11.6}               & \textbf{100.0*} / \textbf{100.0*} & 65.4 / \textbf{86.9}              & 4.4 / \textbf{9.9}                \\
		& MI / +BPFA     & 9.9 / \textbf{36.5}               & \textbf{100.0*} / \textbf{100.0*} & 74.0 / \textbf{94.0}              & 7.1 / \textbf{15.1}               \\
		& DI / +BPFA     & 32.5 / \textbf{36.6}              & \textbf{100.0*} / \textbf{100.0*} & 97.3 / \textbf{98.3}              & 17.7 / \textbf{22.8}              \\
		& DFANet / +BPFA & 8.2 / \textbf{11.0}               & \textbf{100.0*} / \textbf{100.0*} & 79.0 / \textbf{86.6}              & 5.8 / \textbf{9.1}                \\
		& SSA / +BPFA    & 12.1 / \textbf{20.7}              & \textbf{100.0*} / \textbf{100.0*} & 90.3 / \textbf{96.4}              & 8.0 / \textbf{18.1}               \\ \hline
		\multirow{5}{*}{IRSE50}  & FIM / +BPFA    & 12.7 / \textbf{23.5}              & 74.4 / \textbf{91.6}              & \textbf{100.0*} / \textbf{100.0*} & 27.3 / \textbf{50.3}              \\
		& MI / +BPFA     & 16.3 / \textbf{30.8}              & 73.2 / \textbf{95.6}              & \textbf{100.0*} / \textbf{100.0*} & 30.3 / \textbf{57.8}              \\
		& DI / +BPFA     & 45.6 / \textbf{51.5}              & 97.7 / \textbf{98.4}              & \textbf{100.0*} / \textbf{100.0*} & 57.5 / \textbf{65.8}              \\
		& DFANet / +BPFA & 20.9 / \textbf{28.1}              & 93.5 / \textbf{96.2}              & \textbf{100.0*} / \textbf{100.0*} & 37.5 / \textbf{51.8}              \\
		& SSA / +BPFA    & 38.6 / \textbf{41.1}              & 97.5 / \textbf{97.7}              & \textbf{100.0*} / \textbf{100.0*} & 59.1 / \textbf{62.5}              \\ \hline
		\multirow{5}{*}{IR152}   & FIM / +BPFA    & 8.8 / \textbf{12.7}               & 5.2 / \textbf{9.4}                & 28.8 / \textbf{51.1}              & \textbf{100.0*} / \textbf{100.0*} \\
		& MI / +BPFA     & 11.5 / \textbf{22.3}              & 6.5 / \textbf{16.4}               & 27.1 / \textbf{57.9}              & \textbf{100.0*} / \textbf{100.0*} \\
		& DI / +BPFA     & 24.0 / \textbf{26.1}              & 15.2 / \textbf{17.8}              & 49.7 / \textbf{57.9}                  & \textbf{100.0*} / \textbf{100.0*} \\
		& DFANet / +BPFA & 13.5 / \textbf{17.3}              & 10.2 / \textbf{14.3}              & 50.6 / \textbf{54.9}              & \textbf{100.0*} / \textbf{100.0*} \\
		& SSA / +BPFA    & 18.9 / \textbf{22.7}              & 18.0 / \textbf{22.9}              & 54.9 / \textbf{59.1}              & \textbf{100.0*} / \textbf{100.0*} \\ \hline
	\end{tabular}
\end{table*}

\begin{table*}[ht]
	\centering
	\small
	\caption{The attack success rates of impersonation attacks on CelebA-HQ with the state-of-the-art attacks as the baseline. The first column represents the attacker models. The third to sixth columns in the first row represent the victim models. $*$ indicates white-box attacks.}
	\label{tab:asr_single_attack_celeba_hq}
	\begin{tabular}{c|c|c|c|c|c}
		\hline
		\multicolumn{1}{l|}{}    & Attack         & \multicolumn{1}{c|}{FaceNet}      & \multicolumn{1}{c|}{MF}           & \multicolumn{1}{c|}{IRSE50}       & \multicolumn{1}{c}{IR152}         \\ \hline
		\multirow{5}{*}{FaceNet} & FIM / +BPFA    & \textbf{100.0*} / \textbf{100.0*} & 8.6 / \textbf{17.3}               & 14.8 / \textbf{26.5}              & 9.2 / \textbf{15.7}               \\
		& MI / +BPFA     & \textbf{100.0*} / \textbf{100.0*} & 11.6 / \textbf{25.1}              & 21.0 / \textbf{36.5}              & 17.1 / \textbf{24.4}              \\
		& DI / +BPFA     & \textbf{100.0*} / \textbf{100.0*} & 22.4 / \textbf{26.0}              & 30.3 / \textbf{37.2}              & 21.0 / \textbf{26.0}              \\
		& DFANet / +BPFA & \textbf{100.0*} / \textbf{100.0*} & 15.9 / \textbf{19.1}              & 21.2 / \textbf{26.0}              & 12.9 / \textbf{16.8}              \\
		& SSA / +BPFA    & \textbf{100.0*} / \textbf{100.0*} & 32.1 / \textbf{36.2}              & 40.1 / \textbf{44.2}              & 21.9 / \textbf{25.6}              \\ \hline
		\multirow{5}{*}{MF}      & FIM / +BPFA    & 6.9 / \textbf{12.5}               & \textbf{100.0*} / \textbf{100.0*} & 60.6 / \textbf{83.9}              & 6.5 / \textbf{10.6}               \\
		& MI / +BPFA     & 10.6 / \textbf{36.5}              & \textbf{100.0*} / \textbf{100.0*} & 65.2 / \textbf{94.0}              & 9.2 / \textbf{15.1}               \\
		& DI / +BPFA     & 31.3 / \textbf{33.4}              & \textbf{100.0*} / \textbf{100.0*} & 96.6 / \textbf{97.5}              & 24.8 / \textbf{25.7}              \\
		& DFANet / +BPFA & 9.7 / \textbf{10.8}               & \textbf{100.0*} / \textbf{100.0*} & 74.7 / \textbf{81.5}              & 8.3 / \textbf{9.1}                \\
		& SSA / +BPFA    & 20.6 / \textbf{20.7}              & \textbf{100.0*} / \textbf{100.0*} & 95.4 / \textbf{95.7}              & 20.1 / \textbf{21.6}              \\ \hline
		\multirow{5}{*}{IRSE50}  & FIM / +BPFA    & 14.5 / \textbf{23.2}              & 76.2 / \textbf{91.8}              & \textbf{100.0*} / \textbf{100.0*} & 31.8 / \textbf{48.9}              \\
		& MI / +BPFA     & 16.3 / \textbf{28.8}              & 73.6 / \textbf{95.6}              & \textbf{100.0*} / \textbf{100.0*} & 33.2 / \textbf{56.1}              \\
		& DI / +BPFA     & 42.2 / \textbf{44.7}              & 97.5 / \textbf{97.7}     & \textbf{100.0*} / \textbf{100.0*} & 56.7 / \textbf{63.8}              \\
		& DFANet / +BPFA & 21.3 / \textbf{26.6}              & 92.7 / \textbf{94.2}              & \textbf{100.0*} / \textbf{100.0*} & 39.1 / \textbf{48.3}              \\
		& SSA / +BPFA    & 34.1 / \textbf{34.5}              & 97.3 / \textbf{97.6}              & \textbf{100.0*} / \textbf{100.0*} & 58.0 / \textbf{59.6}              \\ \hline
		\multirow{5}{*}{IR152}   & FIM / +BPFA    & 12.4 / \textbf{15.1}              & 11.3 / \textbf{17.0}              & 37.6 / \textbf{56.0}              & \textbf{100.0*} / \textbf{100.0*} \\
		& MI / +BPFA     & 16.1 / \textbf{22.3}              & 14.7 / \textbf{23.8}              & 36.0 / \textbf{62.1}              & \textbf{100.0*} / \textbf{100.0*} \\
		& DI / +BPFA     & 28.0 / \textbf{29.7}              & 26.4 / \textbf{27.1}              & 58.6 / \textbf{62.9}              & \textbf{100.0*} / \textbf{100.0*} \\
		& DFANet / +BPFA & 17.8 / \textbf{21.0}              & 20.7 / \textbf{23.2}              & 56.6 / \textbf{61.1}              & \textbf{100.0*} / \textbf{100.0*} \\
		& SSA / +BPFA    & 25.5 / \textbf{26.0}              & 32.4 / \textbf{33.8}              & 63.9 / \textbf{64.3}              & \textbf{100.0*} / \textbf{100.0*} \\ \hline
	\end{tabular}
\end{table*}
\section{Experiments}
In this section, we present extensive experiments to evaluate the effectiveness of our proposed BPFA and to highlight its properties. As FR technology continues to be used in an increasing number of applications, from access control to financial services, concerns around the need for governance mechanisms to ensure its responsible use have become increasingly important within the digital nation. The experiments presented in this section highlight the potential for our proposed BPFA to provide a more secure and reliable FR system, with improved performance against state-of-the-art attacks. Specifically, Section IV-A introduces the experimental setting. Section \ref{experimental_setting} introduces the experimental setting. Section \ref{comparison_study} presents the experiment results of the proposed BPFA with the state-of-the-art attacks as the baseline. Section \ref{comparison_study_combinations} demonstrates the performance of the proposed BPFA with the attack combinations as the baseline. Section \ref{attack_adv_robust_models} shows the experiment results of the proposed BPFA on adversarial robust models.
Section \ref{dodging_attack} presents the performance of BPFA on the dodging attack setting. Section \ref{ablation_studies} performs ablation experiments on the proposed BPFA. Section \ref{effect_of_adding_beneficial} studies the effectiveness of simulated gradients. Section \ref{interpretation_of_beneficial_perturbations} performs experiments on the interpretation of beneficial perturbations added on the feature maps. Section \ref{storage_cost_of_bpfa} studies the storage overhead of BPFA.
\subsection{Experimental Setting}\label{experimental_setting}
\textbf{Datasets.} We choose LFW\cite{lfw} and CelebA-HQ\cite{celeba_hq} as the datasets for our experiments. LFW is a face dataset for unconstrained FR. CelebA-HQ is a dataset with high-quality images. These two datasets are commonly used in the research of adversarial attacks on FR\cite{genap}\cite{adv_makeup}\cite{amt_gan}.

Following \cite{dfanet}\cite{tip_im}\cite{amt_gan}\cite{genap}\cite{evolutionary}, we also select a part of the data from the selected dataset as the dataset to test the performance of the attack methods. Specifically, we randomly select 1000 face pairs from the LFW dataset as the dataset for LFW and 1000 face pairs from the CelebA-HQ dataset as the dataset for CelebA-HQ. The number of face pairs we selected is comparable to the previous works. All the pairs we selected are negative pairs, where one image in a pair is used as the attacker image and the other as the victim image. Without particular emphasis, we will use LFW and CelebA-HQ to refer to our selected LFW and CelebA-HQ datasets, respectively.

\textbf{Face Recognition Models.} The FR models used in our experiments are FaceNet, MobileFace (we denote it as MF in the following), IRSE50, and IR152 which are widely used in various applications such as surveillance, access control, and financial services in the digital nation. However, the increasing use of FR technology also raises concerns about the need for governance mechanisms to ensure its responsibility. We choose the thresholds when FAR@0.001 and FAR@0.01 on the whole LFW dataset as the thresholds to calculate the attack success rate for the impersonation attacks and dodging attacks, respectively.

\textbf{Attack Setting.}
Most of the attacks in the experiments are conducted under the setting of impersonation attacks because impersonation attacks are more difficult compared to dodging attacks\cite{dfanet}\cite{liu_delving_into}.
The findings from these experiments are particularly relevant to the broader issue of governance within the digital nation, given the extensive use of FR in sensitive scenarios. To ensure the responsible use of FR in these sensitive scenarios, it is crucial to understand the potential vulnerabilities of such systems to impersonation attacks. By conducting experiments and evaluating the robustness of facial recognition models against impersonation attacks, our study contributes to advancing the ongoing conversation around governance mechanisms for FR within the digital nation.
Following \cite{dfanet}, we set the maximum allowable perturbation magnitude $\epsilon$ to 10 based on $L_{\infty}$ norm bound with respect to pixel values in $[0, 255]$, and the maximum iterative step to 1500. For all the attack methods, we set the step size $\beta$ to 1.0.
The locations where BPFA adds beneficial perturbations are on the feature maps of the convolutional layers if not specified.

\textbf{Evaluation Metrics.} We use \textit{attack success rate} (ASR) to evaluate the performance of different attacks. ASR is the ratio of adversarial face examples that attack successfully.

Because the objectives of the impersonation and dodging attacks are different, the methods for calculating the ASR for the two types of attacks differ.
For impersonation attacks on FR, ASR can be calculated using the following formula:
\begin{equation}
	{\rm ASR}^i=\frac{\sum_{i=1}^{N_p}\mathds{1}\left(\mathcal{D}\left(\mathcal{F}^{vct}\left(x^{adv}\right), \mathcal{F}^{vct}\left(x^{t}\right)\right)<t^i\right)}{N_p}\label{eq:asr_impersonation}
\end{equation}
where $N_p$ is the number of face pairs, $t^i$ is the threshold for impersonation attacks.
For dodging attacks on FR, ASR can be calculated using the following formula:
\begin{equation}
	{\rm ASR}^d=\frac{\sum_{i=1}^{N_p}\mathds{1}\left(\mathcal{D}\left(\mathcal{F}^{vct}\left(x^{adv}\right), \mathcal{F}^{vct}\left(x^{s}\right)\right)>t^d\right)}{N_p}\label{eq:asr_dodging}
\end{equation}
where $t^d$ is the threshold for dodging attacks.

\textbf{Baseline methods} Since Adv-Makeup\cite{adv_makeup} and AMT-GAN\cite{amt_gan} are attacks based on makeup transfer and GenAP is a patch-based attack, it is not rational to use $L_{\infty}$ norm to limit their maximum allowable perturbation magnitude on the whole face images. TIP-IM\cite{tip_im} is an attack used for face encryption that requires high visual quality. Though TIP-IM brings a significant improvement to the performance of face encryption, our experiments demonstrate that if TIP-IM is directly applied under the setting of this paper, the ASR of the combination of FIM\cite{fim}, MI\cite{mim}, DI\cite{dim}, and TIP-IM is lower than the combination of FIM, MI, DI which we denote it as FMD. Therefore, Adv-Makeup, AMT-GAN, GenAP, and TIP-IM are not used as baselines in this paper.

For the setting of this paper, the state-of-the-art attack method on FR is the combination of FIM\cite{fim}, MI\cite{mim}, DI\cite{dim}, and DFANet\cite{dfanet} (we denote it as FMDN) which is shown in the literature\cite{dfanet}. Therefore, we mainly use FMDN as our baseline. FR and image classification are both computer vision tasks. As shown in \cite{amt_gan}, the advanced transferable adversarial attacks on image classification may also be effective on FR. To show the effectiveness of our proposed BPFA, we also plan to use diverse state-of-the-art transferable adversarial attacks and attack combinations on image classification as our baseline. Specifically, the baseline which we choose includes MI\cite{mim}, DI\cite{dim}, SSA\cite{ssa6}, SNI (the combination of SI\cite{ni_fgsm} and NI\cite{ni_fgsm}), and VMI (the combination of VT\cite{vt} and MI).
\subsection{Comparison Study with the State-Of-The-Art Attacks}\label{comparison_study}
To evaluate the effectiveness of our proposed BPFA, we test the performance of BPFA with the state-of-the-art attacks on the LFW and CelebA-HQ datasets.

The results of different attacks on LFW and CelebA-HQ are shown in Table \ref{tab:asr_single_attack_lfw} and Table \ref{tab:asr_single_attack_celeba_hq}, respectively.
In Table \ref{tab:asr_single_attack_lfw} and Table \ref{tab:asr_single_attack_celeba_hq}, the number before / is the ASR of the baseline method, and the number after / is the ASR of the baseline method after adding BPFA, which is same in Table \ref{tab:asr_attack_combination_lfw}, Table \ref{tab:asr_attack_combination_celeba_hq}, Table \ref{tab:asr_attack_at_model_lfw}, and Table \ref{tab:asr_attack_at_model_celeba_hq}.
Table \ref{tab:asr_single_attack_lfw} and Table \ref{tab:asr_single_attack_celeba_hq} show that the attacks, when supplemented with BPFA, exhibit superior black-box performance compared to the attacks without the addition of BPFA. This demonstrates that BPFA can be well combined with existing adversarial attacks to boost the black-box performance of them. While this finding has potential implications for improving the adversarial robustness of FR systems, it also underscores the need for governance mechanisms that ensure the responsible use of such systems in the digital nation. By demonstrating the effectiveness of BPFA to enhance the black-box performance of existing adversarial attacks on FR, our study highlights the need for governance mechanisms that can help detect and mitigate adversarial attacks on FR in the digital nation. Such governance mechanisms can facilitate the use of FR systems within the digital nation in a responsible and trustworthy way, ensuring their effective deployment for the benefit of society.

\subsection{Comparison Study with the Attack Combinations}\label{comparison_study_combinations}
In practice, attackers can combine multiple attack methods against the FR systems in the digital nation to achieve better attack performance. Therefore, it is crucial to evaluate whether BPFA is effective in improving the performance of attack combinations to ensure that governance strategies are successful in protecting against the practical adversarial attacks on FR. To this end, we test the performance of BPFA with the attack combinations on the LFW and CelebA-HQ datasets.

The results on LFW and CelebA-HQ are demonstrated in Table \ref{tab:asr_attack_combination_lfw} and Table \ref{tab:asr_attack_combination_celeba_hq}, respectively.
Table \ref{tab:asr_attack_combination_lfw} and Table \ref{tab:asr_attack_combination_celeba_hq} provide evidence that BPFA enhances the transferability of attack combinations, thereby bolstering the effectiveness of our proposed BPFA.

\begin{figure}[htbp]
	\centering
	\includegraphics[width=80mm]{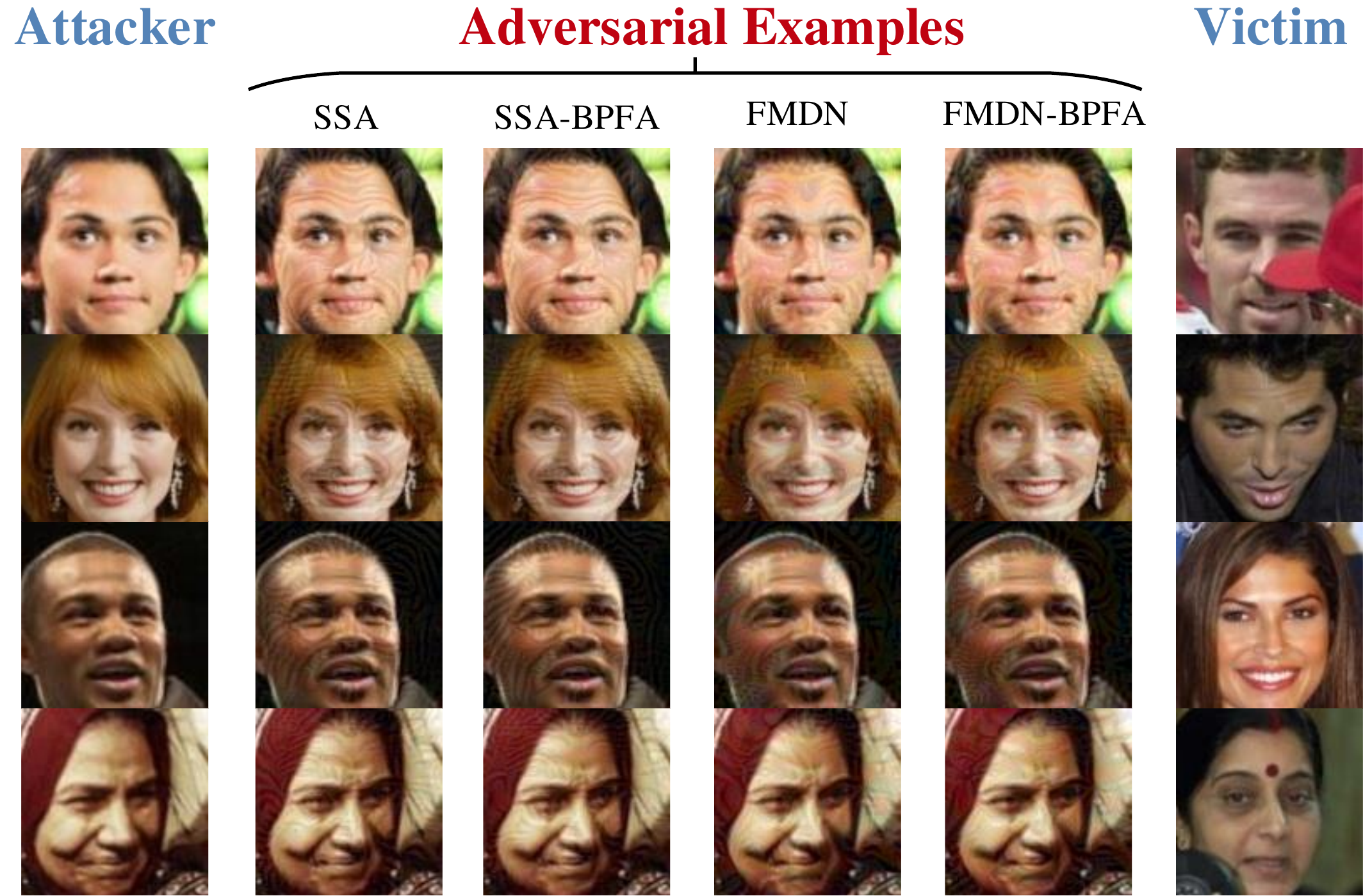}
	\caption{Illustration of the adversarial examples crafted by SSA, SSA-BPFA, FMDN, and FMDN-BPFA. The images in the first and last columns are the attacker images and victim images, respectively. The second to fifth columns demonstrate the adversarial examples crafted by SSA, SSA-BPFA, FMDN, and FMDN-BPFA, respectively.}
	\label{fig:ax_crafted_by_diff_methods_show}
\end{figure}

Examples of adversarial examples crafted by SSA, SSA-BPFA, FMDN, and FMDN-BPFA are illustrated in Fig. \ref{fig:ax_crafted_by_diff_methods_show}.

\begin{table*}[]
	\centering
	\small
	\caption{The attack success rates of impersonation attacks on LFW with the attack combinations as the baseline. The first column represents the attacker models. The third to sixth columns in the first row represent the victim models. $*$ indicates white-box attacks.}
	\label{tab:asr_attack_combination_lfw}
	\begin{tabular}{c|c|c|c|c|c}
		\hline
		& Attack       & FaceNet                  & MF                       & IRSE50                   & IR152                    \\ \hline
		\multirow{4}{*}{FaceNet} & FMD / +BPFA  & \textbf{100.0*} / \textbf{100.0*} & 25.0 / \textbf{32.4}              & 42.9 / \textbf{54.0}              & 29.9 / \textbf{39.3}              \\
		& FMDN / +BPFA & \textbf{100.0*} / \textbf{100.0*} & 30.7 / \textbf{31.8}              & 47.9 / \textbf{52.5}              & 34.5 / \textbf{38.5}              \\
		& SNI / +BPFA  & \textbf{100.0*} / \textbf{100.0*} & 16.5 / \textbf{23.8}              & 31.0 / \textbf{41.4}                & 13.9 / \textbf{25.7}              \\
		& VMI / +BPFA  & \textbf{100.0*} / \textbf{100.0*} & 23.2 / \textbf{27.1}              & 36.3 / \textbf{44.0}              & 26.1 / \textbf{29.8}              \\ \hline
		\multirow{4}{*}{MF}      & FMD / +BPFA  & 40.5 / \textbf{44.5}              & \textbf{100.0*} / \textbf{100.0*} & 98.5 / \textbf{99.1}              & 22.0 / \textbf{27.5}              \\
		& FMDN / +BPFA & 40.8 / \textbf{45.4}              & \textbf{100.0*} / \textbf{100.0*} & 98.8 / \textbf{99.0}              & 23.5 / \textbf{27.1}              \\
		& SNI / +BPFA  & 16.9 / \textbf{19.3}              & \textbf{100.0*} / \textbf{100.0*} & 94.4 / \textbf{96.1}              & 11.7  / \textbf{15.4}             \\
		& VMI / +BPFA  & 12.2 / \textbf{21.9}              & \textbf{100.0*} / \textbf{100.0*} & 90.5 / \textbf{96.0}              & 8.6 / \textbf{17.3}               \\ \hline
		\multirow{4}{*}{IRSE50}  & FMD / +BPFA  & 55.1 / \textbf{57.9}              & 98.8 / \textbf{99.0}              & \textbf{100.0*} / \textbf{100.0*} & 64.8 / \textbf{71.2}              \\
		& FMDN / +BPFA & 58.5 / \textbf{59.3}              & 99.0 / \textbf{99.1}              & \textbf{100.0*} / \textbf{100.0*} & 68.2 / \textbf{71.5}              \\
		& SNI / +BPFA  & 26.2 / \textbf{37.0}              & 91.6 / \textbf{96.5}              & \textbf{100.0*} / \textbf{100.0*} & 46.1 / \textbf{61.9}              \\
		& VMI / +BPFA  & 23.3 / \textbf{36.0}              & 90.3 / \textbf{96.9}              & \textbf{100.0*} / \textbf{100.0*} & 41.8 / \textbf{62.7}              \\ \hline
		\multirow{4}{*}{IR152}   & FMD / +BPFA  & 32.3 / \textbf{33.9}              & 23.0 / \textbf{25.4}              & 53.0 / \textbf{61.3}              & \textbf{100.0*} / \textbf{100.0*} \\
		& FMDN / +BPFA & 37.7 / \textbf{39.6}              & 28.6 / \textbf{29.6}              & 58.3 / \textbf{64.2}              & \textbf{100.0*} / \textbf{100.0*} \\
		& SNI / +BPFA  & 12.9 / \textbf{18.2}              & 8.5 / \textbf{15.1}               & 34.2 / \textbf{54.2}              & \textbf{100.0*} / \textbf{100.0*} \\
		& VMI / +BPFA  & 15.8 / \textbf{22.0}              & 14.9 / \textbf{21.1}              & 50.5 / \textbf{59.1}              & \textbf{100.0*} / \textbf{100.0*} \\ \hline
	\end{tabular}
\end{table*}

\begin{table*}[]
	\centering
	\small
	\caption{
		The attack success rates of impersonation attacks on CelebA-HQ with the attack combinations as the baseline.
		The first column represents the attacker models. The third to sixth columns in the first row represent the victim models. $*$ indicates white-box attacks.}
	\label{tab:asr_attack_combination_celeba_hq}
	\begin{tabular}{c|c|c|c|c|c}
		\hline
		\multicolumn{1}{l|}{}    & Attack                            & FaceNet                           & MF                                & IRSE50                            & IR152                             \\ \hline
		\multirow{4}{*}{FaceNet} & \multicolumn{1}{c|}{FMD / +BPFA}  & \textbf{100.0*} / \textbf{100.0*} & 27.8 / \textbf{35.0}              & 38.7 / \textbf{46.3}              & 29.1 / \textbf{34.2}              \\
		& \multicolumn{1}{c|}{FMDN / +BPFA} & \textbf{100.0*} / \textbf{100.0*} & 31.6 / \textbf{35.8}              & 42.2 / \textbf{46.4}              & 30.9 / \textbf{34.9}              \\
		& \multicolumn{1}{c|}{SNI / +BPFA}  & \textbf{100.0*} / \textbf{100.0*} & 23.1 / \textbf{30.2}              & 31.6 / \textbf{40.5}              & 18.0 / \textbf{23.8}              \\
		& \multicolumn{1}{c|}{VMI / +BPFA}  & \textbf{100.0*} / \textbf{100.0*} & 13.5 / \textbf{26.7}              & 22.8 / \textbf{37.1}              & 15.7 / \textbf{24.9}              \\ \hline
		\multirow{4}{*}{MF}      & \multicolumn{1}{c|}{FMD / +BPFA}  & 36.8 / \textbf{39.7}              & \textbf{100.0*} / \textbf{100.0*} & 98.1 / \textbf{98.4}              & 28.2 / \textbf{30.4}              \\
		& \multicolumn{1}{c|}{FMDN / +BPFA} & 36.8 / \textbf{38.3}              & \textbf{100.0*} / \textbf{100.0*} & 98.1 / \textbf{98.3}              & 28.1 / \textbf{31.3}              \\
		& \multicolumn{1}{c|}{SNI / +BPFA}  & 18.0 / \textbf{19.9}              & \textbf{100.0*} / \textbf{100.0*} & 93.3 / \textbf{95.7}              & 17.4 / \textbf{23.0}              \\
		& \multicolumn{1}{c|}{VMI / +BPFA}  & 14.5 / \textbf{19.2}              & \textbf{100.0*} / \textbf{100.0*} & 90.0 / \textbf{95.5}              & 12.9 / \textbf{19.8}              \\ \hline
		\multirow{4}{*}{IRSE50}  & \multicolumn{1}{c|}{FMD / +BPFA}  & 48.1 / \textbf{49.3}              & 98.7 / \textbf{98.8}              & \textbf{100.0*} / \textbf{100.0*} & 63.4 / \textbf{64.5}              \\
		& \multicolumn{1}{c|}{FMDN / +BPFA} & 49.3 / \textbf{51.2}              & 98.7 / \textbf{98.8}              & \textbf{100.0*} / \textbf{100.0*} & 63.6 / \textbf{66.9}              \\
		& \multicolumn{1}{c|}{SNI / +BPFA}  & 26.2 / \textbf{37.0}              & 91.8 / \textbf{96.5}              & \textbf{100.0*} / \textbf{100.0*} & 53.7 / \textbf{61.9}              \\
		& \multicolumn{1}{c|}{VMI / +BPFA}  & 24.3 / \textbf{30.4}              & 93.0 / \textbf{96.3}              & \textbf{100.0*} / \textbf{100.0*} & 46.9 / \textbf{58.7}              \\ \hline
		\multirow{4}{*}{IR152}   & \multicolumn{1}{c|}{FMD / +BPFA}  & 36.5 / \textbf{36.8}              & 34.0 / \textbf{35.7}              & 61.5 / \textbf{67.5}              & \textbf{100.0*} / \textbf{100.0*} \\
		& \multicolumn{1}{c|}{FMDN / +BPFA} & 44.0 / \textbf{46.5}              & 43.8 / \textbf{44.4}              & 75.5 / \textbf{77.1}              & \textbf{100.0*} / \textbf{100.0*} \\
		& \multicolumn{1}{c|}{SNI / +BPFA}  & 16.6 / \textbf{20.0}              & 16.9 / \textbf{21.8}              & 43.2 / \textbf{52.9}              & \textbf{100.0*} / \textbf{100.0*} \\
		& \multicolumn{1}{c|}{VMI / +BPFA}  & 20.4 / \textbf{25.8}              & 20.3 / \textbf{26.9}              & 46.1 / \textbf{62.2}              & \textbf{100.0*} / \textbf{100.0*} \\ \hline
	\end{tabular}
\end{table*}

\subsection{The Attack Performance on Adversarial Robust Models}\label{attack_adv_robust_models}
The implementation of an adversarial robust model by victims in the digital nation to improve their defensive abilities against adversarial face examples highlights the importance of proactive governance strategies to mitigate the threats of adversarial attacks on FR. In this context, the evaluation of the effectiveness of BPFA on adversarial robust FR models becomes crucial to assess the robustness and suitability of related governance measures to address evolving adversarial attacks. To this end, we firstly train FaceNet, MF, IRSE50, and IR152 using MTER\cite{fim}, the latest adversarial training method for FR. We denote the adversarial trained models as FaceNet$_{adv}$, MF$_{adv}$, IRSE50$_{adv}$, and IR152$_{adv}$, respectively.
Next, we use the models without adversarial training to craft adversarial face examples and test the performance of the crafted adversarial face examples on these adversarial robust FR models.

The results on LFW and CelebA-HQ are shown in Table \ref{tab:asr_attack_at_model_lfw} and Table \ref{tab:asr_attack_at_model_celeba_hq}, respectively.
Table \ref{tab:asr_attack_at_model_lfw} and Table \ref{tab:asr_attack_at_model_celeba_hq} demonstrate that the attack performance of the baseline attack methods on the adversarial robust models improves after being combined with BPFA, which shows the effectiveness of BPFA on the adversarial robust models.
\begin{table*}[]
	\centering
	\small
	\caption{The attack success rates of impersonation attacks on LFW with adversarial robust models as victim models. The first column represents the attacker models. The third to sixth columns in the first row represent the victim models.}
	\label{tab:asr_attack_at_model_lfw}
	\begin{tabular}{c|c|c|c|c|c}
		\hline
		\multicolumn{1}{l|}{}    & Attack         & FaceNet$_{adv}$              & MF$_{adv}$                   & IRSE50$_{adv}$               & IR152$_{adv}$                \\ \hline
		\multirow{5}{*}{FaceNet} & FIM / +BPFA    & 15.3 / \textbf{20.7} & 2.2 / \textbf{5.2}   & 5.9 / \textbf{10.7}  & 5.6 / \textbf{12.2}  \\
		& MI / +BPFA     & 25.9 / \textbf{30.1} & 6.5 / \textbf{10.5}  & 12.8 / \textbf{18.2} & 14.1 / \textbf{20.8} \\
		& DI / +BPFA     & 30.2 / \textbf{35.1} & 10.1 / \textbf{12.5} & 15.5 / \textbf{20.4} & 17.8 / \textbf{23.6} \\
		& DFANet / +BPFA & 22.6 / \textbf{27.1} & 4.2 / \textbf{7.0}   & 9.3 / \textbf{13.2}  & 9.6 / \textbf{15.3}  \\
		& SSA / +BPFA    & 33.1 / \textbf{34.4} & 10.6 / \textbf{12.6} & 16.2 / \textbf{21.3} & 17.0 / \textbf{21.6} \\ \hline
		\multirow{5}{*}{MF}      & FIM / +BPFA    & 2.2 / \textbf{3.8}   & 9.0 / \textbf{17.2}  & 3.9 / \textbf{7.0}   & 1.2 / \textbf{3.2}   \\
		& MI / +BPFA     & 3.3 / \textbf{4.0}   & 16.5 / \textbf{26.6} & 8.7 / \textbf{12.8}  & 4.0 / \textbf{5.4}   \\
		& DI / +BPFA     & 10.7 / \textbf{11.5} & 36.8 / \textbf{43.9} & 16.9 / \textbf{20.4} & 9.9 / \textbf{11.6}  \\
		& DFANet / +BPFA & 2.8 / \textbf{3.9}   & 12.4 / \textbf{16.3} & 4.7 / \textbf{6.1}   & 2.4 / \textbf{3.9}   \\
		& SSA / +BPFA    & 4.2 / \textbf{6.7}   & 19.2 / \textbf{34.1} & 7.3 / \textbf{12.6}  & 3.3 / \textbf{7.3}   \\ \hline
		\multirow{5}{*}{IRSE50}  & FIM / +BPFA    & 4.1 / \textbf{6.5}   & 8.0 / \textbf{11.7}  & 19.4 / \textbf{28.8} & 9.1 / \textbf{14.5}  \\
		& MI / +BPFA     & 6.0 / \textbf{9.9}   & 6.9 / \textbf{14.9}  & 18.0 / \textbf{33.8} & 11.1 / \textbf{19.7} \\
		& DI / +BPFA     & 15.2 / \textbf{17.4} & 23.0 / \textbf{26.3} & 40.7 / \textbf{46.5} & 26.1 / \textbf{30.1} \\
		& DFANet / +BPFA & 7.1 / \textbf{9.0}   & 8.8 / \textbf{12.4}  & 23.2 / \textbf{31.8} & 13.2 / \textbf{18.3} \\
		& SSA / +BPFA    & 12.8 / \textbf{13.8} & 19.5 / \textbf{20.8} & 40.2 / \textbf{40.7} & 23.3 / \textbf{25.0} \\ \hline
		\multirow{5}{*}{IR152}   & FIM / +BPFA    & 3.6 / \textbf{3.9}   & 1.9 / \textbf{2.7}   & 6.1 / \textbf{8.1}   & 13.2 / \textbf{22.0} \\
		& MI / +BPFA     & 5.8 / \textbf{8.6}   & 3.3 / \textbf{6.8}   & 9.2 / \textbf{14.8}  & 15.8 / \textbf{33.7} \\
		& DI / +BPFA     & 8.3 / \textbf{8.4}   & 5.4 / \textbf{6.0}   & 12.8 / \textbf{15.4} & 32.0 / \textbf{34.1} \\
		& DFANet / +BPFA & 4.8 / \textbf{7.1}   & 2.4 / \textbf{3.1}   & 9.4 / \textbf{10.8}  & 24.6 / \textbf{28.1} \\
		& SSA / +BPFA    & 7.5 / \textbf{8.2}   & 3.7 / \textbf{5.3}   & 12.1 / \textbf{15.2} & 31.3 / \textbf{36.4} \\ \hline
	\end{tabular}
\end{table*}

\begin{table*}[]
	\centering
	\small
	\caption{The attack success rates of impersonation attacks on CelebA-HQ with adversarial robust models as victim models. The first column represents the attacker models. The third to sixth columns in the first row represent the victim models.}
	\label{tab:asr_attack_at_model_celeba_hq}
	\begin{tabular}{c|c|c|c|c|c}
		\hline
		\multicolumn{1}{l|}{}    & Attack         & FaceNet$_{adv}$              & MF$_{adv}$             & IRSE50$_{adv}$       & IR152$_{adv}$        \\ \hline
		\multirow{5}{*}{FaceNet} & FIM / +BPFA    & 11.7 / \textbf{15.0} & 3.4 / \textbf{6.0}   & 7.6 / \textbf{11.0}  & 6.5 / \textbf{9.9}   \\
		& MI / +BPFA     & 20.2 / \textbf{21.1} & 8.1 / \textbf{11.3}  & 16.3 / \textbf{18.0} & 11.9 / \textbf{15.7} \\
		& DI / +BPFA     & 20.8 / \textbf{23.1} & 11.4 / \textbf{13.1} & 18.6 / \textbf{20.6} & 14.5 / \textbf{17.3} \\
		& DFANet / +BPFA & 16.1 / \textbf{18.2} & 5.4 / \textbf{7.5}   & 11.0 / \textbf{14.4} & 8.3 / \textbf{11.2}  \\
		& SSA / +BPFA    & 22.8 / \textbf{24.9} & 11.1 / \textbf{14.1} & 18.6 / \textbf{20.3} & 14.0 / \textbf{15.5} \\ \hline
		\multirow{5}{*}{MF}      & FIM / +BPFA    & 2.0 / \textbf{2.6}   & 10.6 / \textbf{16.7} & 5.0 / \textbf{8.4}   & 2.5 / \textbf{4.0}   \\
		& MI / +BPFA     & 3.3 / \textbf{4.0}   & 16.5 / \textbf{26.6} & 8.7 / \textbf{12.8}  & 4.0 / \textbf{5.4}   \\
		& DI / +BPFA     & 8.0 / \textbf{9.1}   & 37.7 / \textbf{39.9} & 20.5 / \textbf{22.4} & 10.3 / \textbf{11.7} \\
		& DFANet / +BPFA & 2.2 / \textbf{2.5}   & 13.0 / \textbf{16.2} & 6.6 / \textbf{7.4}   & 2.9 / \textbf{3.2}   \\
		& SSA / +BPFA    & 5.2 / \textbf{5.3}   & 32.4 / \textbf{33.4} & 15.8 / \textbf{16.0} & 7.5 / \textbf{8.2}   \\ \hline
		\multirow{5}{*}{IRSE50}  & FIM / +BPFA    & 4.1 / \textbf{6.5}   & 8.0 / \textbf{11.7}  & 19.4 / \textbf{28.8} & 9.1 / \textbf{14.5}  \\
		& MI / +BPFA     & 5.6 / \textbf{7.2}   & 10.4 / \textbf{15.5} & 23.1 / \textbf{35.2} & 10.8 / \textbf{16.7} \\
		& DI / +BPFA     & 11.4 / \textbf{14.2} & 27.1 / \textbf{29.8} & 39.6 / \textbf{42.1} & 19.9 / \textbf{22.5} \\
		& DFANet / +BPFA & 6.2 / \textbf{7.3}   & 12.9 / \textbf{14.7} & 27.4 / \textbf{31.1} & 12.3 / \textbf{15.4} \\
		& SSA / +BPFA    & 9.0 / \textbf{9.4}   & 22.6 / \textbf{23.8} & 35.5 / \textbf{39.0} & 17.4 / \textbf{19.5} \\ \hline
		\multirow{5}{*}{IR152}   & FIM / +BPFA    & 4.0 / \textbf{5.5}   & 4.8 / \textbf{5.1}   & 11.7 / \textbf{12.8} & 17.7 / \textbf{21.4} \\
		& MI / +BPFA     & 5.5 / \textbf{8.6}   & 7.0 / \textbf{9.8}   & 14.0 / \textbf{20.1} & 21.2 / \textbf{30.2} \\
		& DI / +BPFA     & 9.1 / \textbf{9.6}   & 10.3 / \textbf{10.4} & 20.5 / \textbf{21.7} & 31.5 / \textbf{33.8} \\
		& DFANet / +BPFA & 5.6 / \textbf{5.9}   & 5.6 / \textbf{6.8}   & 15.4 / \textbf{17.1} & 24.4 / \textbf{28.5} \\
		& SSA / +BPFA    & 9.2 / \textbf{9.4}   & 9.9 / \textbf{10.6}  & 20.1 / \textbf{21.2} & 33.5 / \textbf{34.1} \\ \hline
	\end{tabular}
\end{table*}

\subsection{The Attack Performance on Dodging Attack Setting}\label{dodging_attack}
Except for the impersonation attacks, the dodging attacks are also important components of adversarial attacks for evaluating the adversarial vulnerability of existing FR models.
Therefore, we also test the performance of our proposed BPFA on the dodging attack setting. Specifically, we craft the adversarial face examples using the FMND and FMND+BPFA methods on LFW with FaceNet as the attacker model. The results of attacking normal models and attacking adversarial robust models are illustrated in Table \ref{tab:dodging_attack_normal_models} and Table \ref{tab:dodging_attack_defensive_models}, respectively.

\begin{table}[]
	\centering
	\small
	\caption{The attack success rates of dodging attacks on LFW with normal models as victim models.
		The first column represents the attack methods.
		The second to fifth columns in the first row represent the victim models. $*$ indicates white-box attacks.}
	\label{tab:dodging_attack_normal_models}
	\begin{tabular}{c|c|c|c|c}
		\hline
		Attack    & FaceNet       & MF            & IRSE50        & IR152         \\ \hline
		FMND      & \textbf{100*} & 45.8          & 29.2          & 41.6          \\
		FMND+BPFA & \textbf{100*} & \textbf{51.6} & \textbf{40.1} & \textbf{57.1} \\ \hline
	\end{tabular}
\end{table}

\begin{table}[]
	\centering
	\small
	\caption{The attack success rates of dodging attacks on LFW with adversarial robust models as victim models.
		The first column represents the attack methods.
		The second to fifth columns in the first row represent the victim models.}
	\label{tab:dodging_attack_defensive_models}
	\begin{tabular}{c|c|c|c|c}
		\hline
		Attack    & FaceNet$_{adv}$       & MF$_{adv}$            & IRSE50$_{adv}$        & IR152$_{adv}$         \\ \hline
		FMND      & 89.2 & 7.5          & 12.2          & 48.5          \\
		FMND+BPFA & \textbf{93.3} & \textbf{12.9} & \textbf{24.7} & \textbf{70.4} \\ \hline
	\end{tabular}
\end{table}

Table \ref{tab:dodging_attack_normal_models} and Table \ref{tab:dodging_attack_defensive_models} demonstrate the effectiveness of BPFA in improving the transferability of adversarial face examples in the dodging attack setting. The dodging attack technique can be employed by criminal suspects to evade detection by law enforcement agencies of the government within the digital nation. Hence, the effectiveness of BPFA in mitigating dodging attacks further highlights the need for robust governance frameworks to ensure the safety and security of the digital nation.
\subsection{Ablation Studies}\label{ablation_studies}
We conduct ablation experiments to study the impact of hyperparameters, including the step size $\eta$ of the beneficial perturbations, the layers where the beneficial perturbations are added on, and the number of iterations.
\subsubsection{Effect of the Step Size of Beneficial Perturbations}
\begin{figure}[htbp]
	\centering
	\includegraphics[width=80mm]{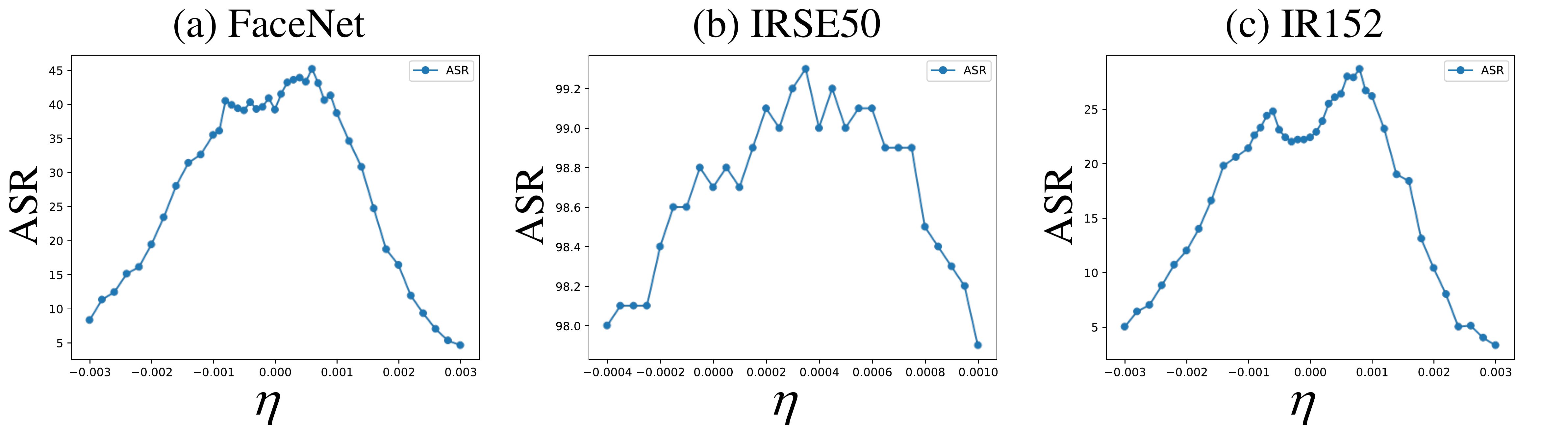}
	\caption{ASR of FMDN-BPFA under different $\eta$ with MF as the attacker model on the LFW dataset under the black-box attack setting.}
	\label{fig:asr_on_diff_eta}
\end{figure}

We test the black-box attack performance of FMDN-BPFA on the impersonation attack setting under different $\eta$ on the LFW dataset with MF as the attacker model, where the beneficial perturbations were added on the feature maps of all convolutional layers. The experiment results are illustrated in  Fig. \ref{fig:asr_on_diff_eta}. 

Fig. \ref{fig:asr_on_diff_eta} shows that with the increase of $\eta$, the black-box attack performance of the adversarial examples crafted by FMDN-BPFA generally shows a trend of first increasing and then decreasing. When $\eta$ is 0, FMDN-BPFA degenerates to FMDN. When $\eta$ is less than 0, the attack method adds adversarial perturbations instead of beneficial perturbations on the feature maps. To analyze the reasons behind Fig. \ref{fig:asr_on_diff_eta}, we should consider how BPFA works. BPFA adds beneficial perturbations on the feature maps to be pitted against the optimization process of adversarial perturbations added on the input image to increase the loss, thereby improving the transferability of the adversarial examples. When $\eta$ is greater than 0, if the value of $\eta$ is too small, the strength of the beneficial perturbations added on the feature maps can not be pitted against the adversarial examples well. If the value of $\eta$ is too large, the important feature will be destroyed by the beneficial perturbations. Therefore, the curve first increases and then decreases when $\eta$ is greater than 0.
When $\eta$ is less than 0, we add adversarial perturbations on the feature maps. We name this attack method as \textbf{A}dversarial \textbf{P}erturbation \textbf{F}eature Augmentation \textbf{A}ttack (APFA). Note that APFA can also improve the transferability of the crafted adversarial examples in some cases. We argue that the improvement of the transferability is due to the important information contained in the adversarial perturbations crafted by gradients (see Section \ref{interpretation_of_beneficial_perturbations}). However, Fig. \ref{fig:asr_on_diff_eta} shows that the performance improvement brought by APFA is much lower than BPFA.
\subsubsection{Effect of the Layers for Adding Beneficial Perturbations on}
To conduct studies on the performance of beneficial perturbations added on different layers, we first count the number of different types of layers in FaceNet, MF, IRSE50, and IR152. From the results, we find that the convolutional layers, Batch Normalization layers, and activation layers are the majority of all layers in these models. Therefore, we conduct our experiments on these layers. Specifically, in the process of crafting adversarial examples using FMDN-BPFA, we only add beneficial perturbations on the convolutional layers, Batch Normalization layers, and activation layers, respectively. The experimental results are illustrated in Table \ref{tab:asr_in_diff_layers}.
\begin{table}[]
	\centering
	\small
	\caption{The ASR of BPFA when beneficial perturbations are added on different layers under the impersonation attack setting. The models in the first column represent the attacker models. The third to sixth columns in the first row represent the victim models. \textit{conv}, \textit{bn}, and \textit{act} in the second column represent the convolutional layers, Batch Normalization layers, and activation layers, respectively. $*$ indicates white-box attacks.}
	\label{tab:asr_in_diff_layers}
	\begin{tabular}{c|c|c|c|c|c}
		\hline
		\multicolumn{1}{l|}{}    & layer & FaceNet         & MF              & IRSE50          & IR152           \\ \hline
		\multirow{3}{*}{FaceNet} & \textit{conv}  & \textbf{100.0*} & 34.7            & 53.9            & 39.1            \\
		& \textit{bn}    & \textbf{100.0*} & \textbf{35.3}   & \textbf{54.1}   & \textbf{39.5}   \\
		& \textit{act}   & \textbf{100.0*} & 32.7            & 52.9            & 37.3            \\ \hline
		\multirow{3}{*}{MF}      & \textit{conv}  & \textbf{45.4}   & \textbf{100.0*} & 99              & 27.1            \\
		& \textit{bn}    & 44.4            & \textbf{100.0*} & \textbf{99.1}   & \textbf{28.4}   \\
		& \textit{act}   & 44.5            & \textbf{100.0*} & \textbf{99.1}   & 27              \\ \hline
		\multirow{3}{*}{IRSE50}  & \textit{conv}  & 59.3            & \textbf{99.1}   & \textbf{100.0*} & 71.5            \\
		& \textit{bn}    & 59.1            & \textbf{99.1}   & \textbf{100.0*} & \textbf{72.2}   \\
		& \textit{act}   & \textbf{59.7}   & \textbf{99.1}   & \textbf{100.0*} & 71.2            \\ \hline
		\multirow{3}{*}{IR152}   & \textit{conv}  & \textbf{39.6}   & \textbf{29.6}   & \textbf{64.2}   & \textbf{100.0*} \\
		& \textit{bn}    & 38.3            & 28.8            & 63.5            & \textbf{100.0*} \\
		& \textit{act}   & 38.4            & 27.7            & 63.2            & \textbf{100.0*} \\ \hline
	\end{tabular}
\end{table}

Table \ref{tab:asr_in_diff_layers} demonstrates the effectiveness of our proposed BPFA method since BPFA method is effective when beneficial perturbations are added on different layers.
Table \ref{tab:asr_in_diff_layers} shows that the performance of FMDN-DFANet is only slightly different whether the beneficial perturbations are added on the convolutional layers, Batch Normalization layers, or activation layers. To figure out the reason for this, we should consider the architectures of the attacker models. In the attacker models, the convolutional layers, Batch Normalization layers, and activation layers tend to be next to each other. When beneficial perturbations are added on these layers, they have similar effects on the entire model due to the strong linearity of single-layer networks. Therefore, the performance of FMDN-BPFA with beneficial perturbations added on the convolutional layers, Batch Normalization layers, and activation layers is similar.
\subsubsection{Effect of the Number of Iterations}
The number of iterations used in crafting adversarial examples will have an impact on the attack performance. In order to further understand BPFA, we used FaceNet as the attacker model to test the black-box performance of the FMDN and FMDN-BPFA under different iterations on the LFW dataset.
\begin{figure}[htbp]
	\centering
	\includegraphics[width=80mm]{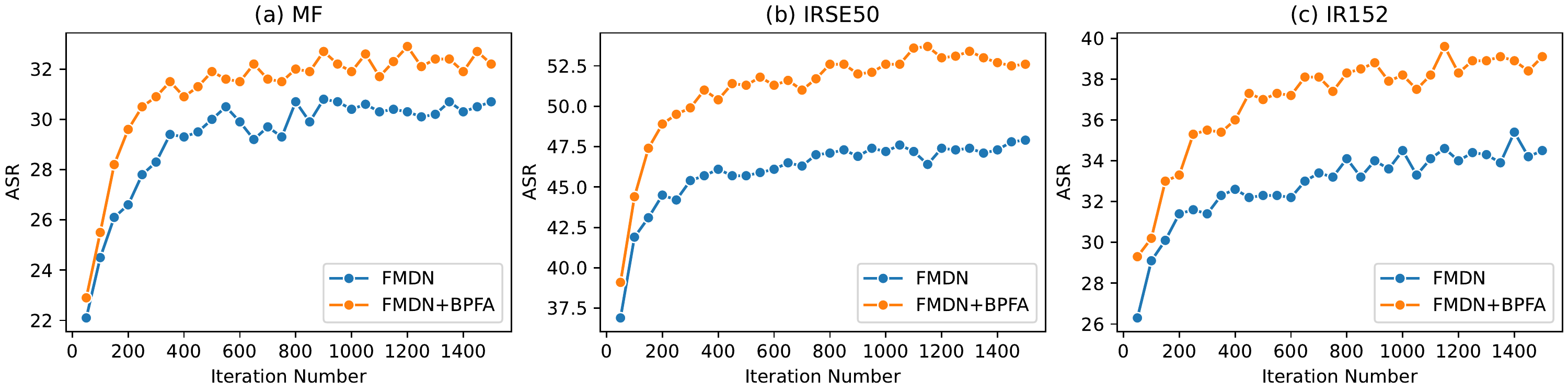}
	\caption{The black-box performance of FMDN and FMDN-BPFA under different iterations with FaceNet as the attacker model on the LFW dataset under the impersonation attack setting.}
	\label{fig:asr_in_diff_round}
\end{figure}

The performance results are shown in Fig. \ref{fig:asr_in_diff_round}. Fig. \ref{fig:asr_in_diff_round} demonstrates that the ASR for both FMDN and FMDN-BPFA increase as the maximum number of iterations $N_{max}$ increase. The gain of ASR of FMDN-BPFA with respect to FMDN increases as $N_{max}$ increases at the beginning of iterations. At the end of iterations, both FMDN and FMDN-BPFA reach convergence, and the ASR of FMDN-BPFA is much higher than that of FMDN, which further verifies the effectiveness of the BPFA method.
It is not hard to see the reason, BPFA can augment the surrogate model with hard models in every iterations, thereby improving the transferability of adversarial face examples.
\subsection{Evaluation of the Effectiveness of Simulated Gradients}\label{effect_of_adding_beneficial}
\begin{figure}[htbp]
	\centering
	\includegraphics[width=80mm]{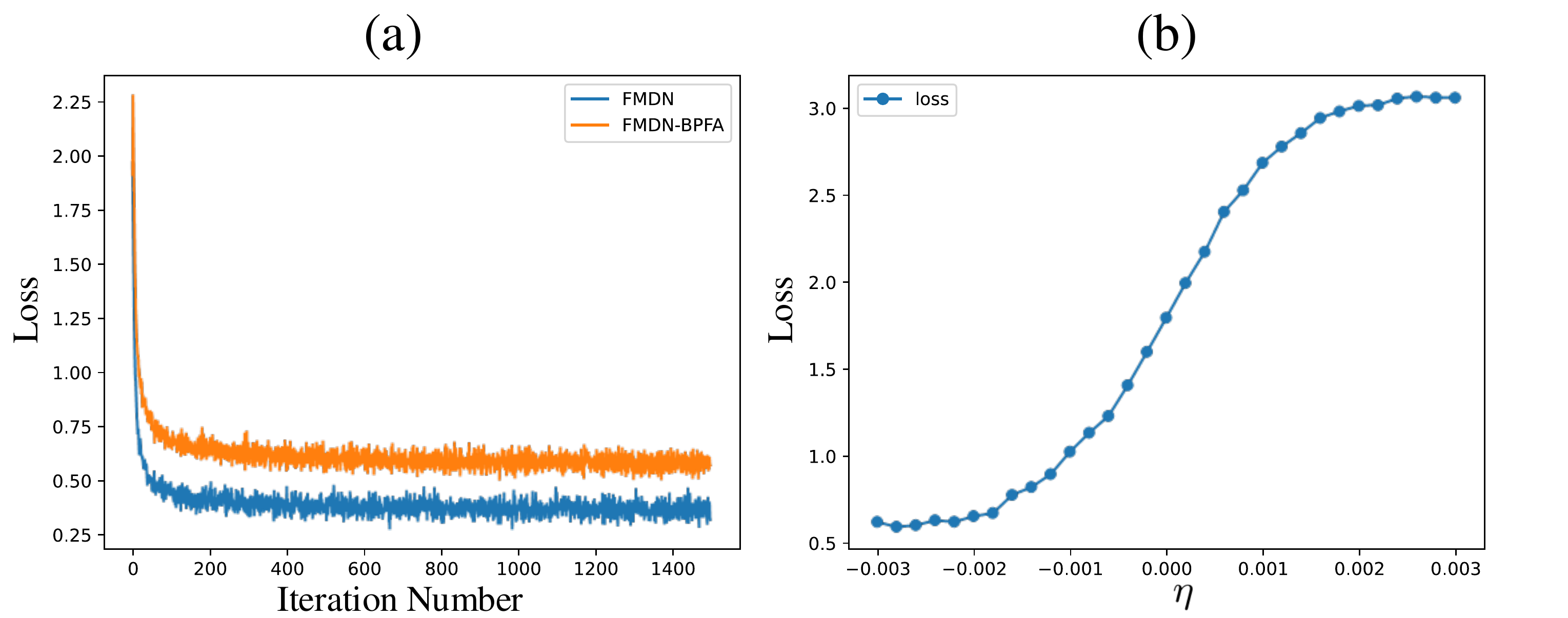}
	\caption{The loss of MF on the LFW dataset under the impersonation attack setting. (a) The loss of FMDN and FMDN-BPFA under different iterations. (b) The loss of FMDN-BPFA under different $\eta$ at the second iteration.}
	\label{fig:loss}
\end{figure}

The purpose of adding beneficial perturbations on the feature maps is to increase the loss to generate hard models. Therefore, we need to verify whether the beneficial perturbations crafted by simulated gradients can increase the loss. To this end, we conduct experiments on the loss of the model with or without adding beneficial perturbations crafted by simulated gradients on the feature maps and the loss of the model after adding beneficial perturbations crafted by simulated gradients under different $\eta$. The experimental results are shown in Fig. \ref{fig:loss}.

Fig. \ref{fig:loss} shows that adding beneficial perturbations crafted by simulated gradients on the feature maps can effectively increase the loss. In addition, subplot (b) of Fig. \ref{fig:loss} demonstrates that the loss increases as $\eta$ increases.
\subsection{Interpretability of Beneficial Perturbations Added on the Feature Maps}\label{interpretation_of_beneficial_perturbations}
\begin{figure}[htbp]
	\centering
	\includegraphics[width=80mm]{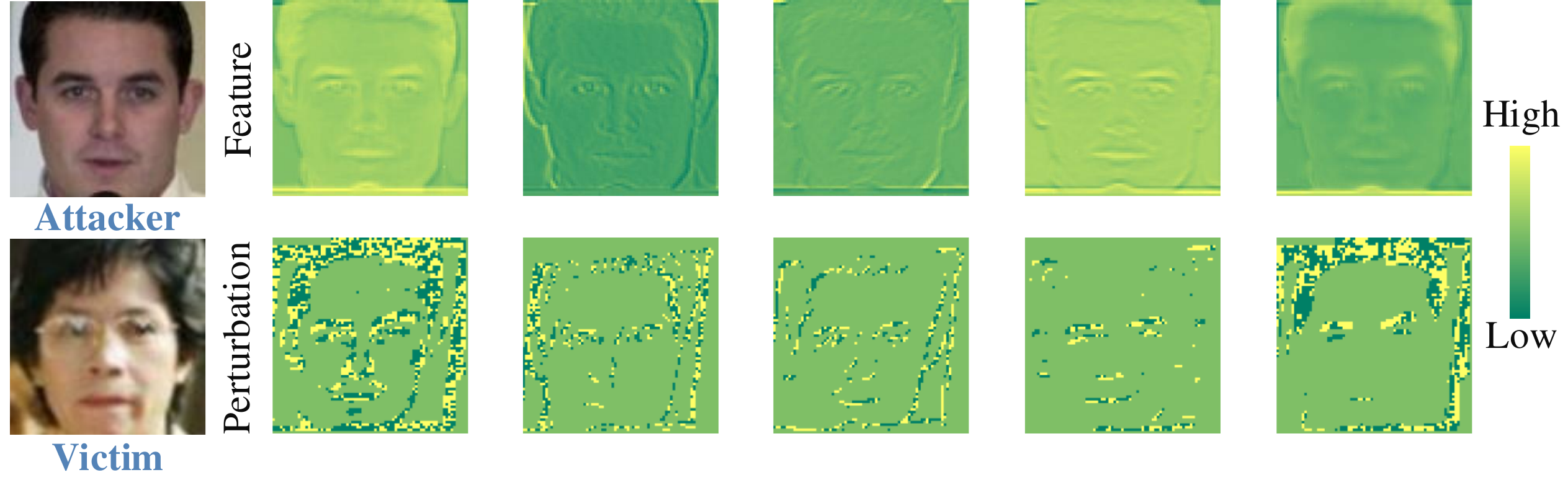}
	\caption{A visualization of the beneficial perturbations added on the feature maps. The two images in the first column are the attacker image and the victim image, respectively. The second to the sixth columns in the first row demonstrate the extracted feature maps, and the second to the sixth columns in the second row demonstrate the beneficial perturbations added on the corresponding feature maps above them.}
	\label{fig:interpretability_bax}
\end{figure}

Our experiments show that adding beneficial perturbations on feature maps to generate hard models can improve the transferability of the crafted adversarial examples. However, the exact form of beneficial perturbations is unknown to us. To address this issue, we study the form of beneficial perturbations added on feature maps to improve the interpretability of them. Specifically, we use FaceNet as the attacker model to craft adversarial examples using the FMDN-BPFA method under the impersonation attack setting. At the second iteration, we visualize the beneficial perturbation added on the feature maps of conv2d\_2a.conv layer of the model, as shown in Fig. \ref{fig:interpretability_bax} for some channels.
We can see the contours of the person from the images of the beneficial perturbations in Fig. \ref{fig:interpretability_bax}, thus suggesting that the beneficial perturbations may have some semantic information.
To analyze the reason behind this, we should first consider how beneficial perturbations are crafted. Beneficial perturbations are derived by performing a sign operation on the gradients. If an element of a gradient tensor has a nonzero value, the sign operation will give it a value whose absolute value is 1 (corresponding to the dark green and light yellow parts in the beneficial perturbations in Fig. \ref{fig:interpretability_bax}). If an element of a gradient tensor has a value of 0, the absolute value of this element will be 0 after the sign operation (corresponding to the light green parts in the beneficial perturbations in Fig. \ref{fig:interpretability_bax}). Therefore, the value of a beneficial perturbation can be seen as a reaction of the magnitude of the gradient to some extent. According to the attribution methods\cite{ig}\cite{not_just}, the gradient can be used to assign attribution value (sometimes also called ``contribution'') to each feature of a network\cite{towards_better_understanding}. The beneficial perturbations added on the feature maps are similar to the saliency maps\cite{sanity_checks} of the feature maps. We argue that for the FR model, certain regions in the feature maps (e.g., contour features) are important, leading to the beneficial perturbations added on the feature maps having some semantic information. Similarly, adversarial perturbations added on the feature maps can also be seen as saliency maps containing important information.

\subsection{Evaluation of the Storage Overhead of BPFA}\label{storage_cost_of_bpfa}
BPFA uses simulated gradients to save computational overhead, which needs to record the gradients in previous iterations. It is meaningful to evaluate whether the storage overhead of BPFA is much higher than that of the baseline. Therefore, we conduct experiments to evaluate the storage overhead of the attack method with or without adding BPFA. Specifically, we craft three adversarial face examples one by one with a size of $\left(112, 112\right)$ using FIM and FIM-BPFA in different FR models under the impersonation attack setting and evaluate the average GPU memory when crafting the adversarial face examples on an NVIDIA GeForce RTX 3090. The results are illustrated in Table \ref{tab:storage_overhead}.
\begin{table}[]
	\centering
	\small
	\caption{The Storage Overhead of different attacks (MB). FIM-BPFA (conv), FIM-BPFA (bn), FIM-BPFA (act) are the BPFA methods that add beneficial perturbations on all the feature maps of the convolutional layers, Batch Normalization layers, and activation layers, respectively.}
	\label{tab:storage_overhead}
	\begin{tabular}{l|c|c|c|c}
		\hline
		\multicolumn{1}{c|}{Attack}   & FaceNet & MF & IRSE50 & IR152\\ \hline
		FIM                              & 2477    & 2267   & 2699    & 2967   \\
		FIM-BPFA (conv)                  & 2489    & 2277    & 2715    & 2993   \\
		FIM-BPFA (bn)                   & 2485    & 2277     & 2715   & 2993    \\
		FIM-BPFA (act)                  & 2497    & 2276     & 2719   & 2979    \\ \hline
	\end{tabular}
\end{table}

Table \ref{tab:storage_overhead} indicates that after adding BPFA to the baseline, the storage overhead increases slightly, while the black-box attack performance improves a lot. We argue that BPFA is worth adopting in cases where the increased storage overhead is not very important.
\section{Conclusion}
In this paper, we study the improvement of the transferability of adversarial face examples to expose more blind spots of FR systems used in the digital nation. Firstly, we propose the concept of hard models based on the property of hard samples. Utilizing the concept of hard models, we propose BPFA to improve the transferability of adversarial face examples. BPFA adds beneficial perturbations on the pre-selected feature maps of the FR model to obtain the effect of hard model augmentations and alleviate the overfitting of the crafted adversarial examples to the surrogate model. In addition, to save the computational overhead of BPFA, we use the gradients in the last backpropagation to simulate the gradients required for computing the beneficial perturbations. We explore the beneficial perturbations added on the feature maps and find that the beneficial perturbations may contain some semantic information. Extensive experiments show that our proposed BPFA can improve the transferability of adversarial face examples.
Moreover, our work can contribute to and assist the stakeholders that we have identified in achieving the development of governance frameworks in nations at higher levels of digitalization. Specifically, our work can benefit citizens by improving the security and privacy of citizens' personal property via identifying and addressing vulnerabilities in FR systems in financial payment which is beneficial for the building process of the digital economy. For the government, our work can aid in enhancing the security of FRs systems, preventing attackers from bypassing these systems and causing potential harm, thereby supporting the development of the digital government. Our work can also assist technology companies in creating more secure and effective FR systems that are resilient against adversarial attacks targeted at FR systems in the digital nation.

\bibliographystyle{IEEEtran}
\bibliography{bpfa_refer}

\end{document}